**Site-Agnostic 3D Dose Distribution Prediction with Deep Learning Neural Networks**

Maryam Mashayekhi[1], Itzel Ramirez Tapia[1], Anjali Balagopal[1], Xinran Zhong[1], Azar Sadeghnejad Barkousaraie[1], Rafe McBeth[1], Mu-Han Lin[1], Steve Jiang[1], Dan Nguyen[1,*]

[1] Medical Artificial Intelligence and Automation Laboratory, Department of Radiation Oncology, University of Texas Southwestern Medical Center, Dallas, TX 75390, United States of America

* Author to whom any correspondence should be addressed.

E-mail: Dan.Nguyen@UTSouthwestern.edu

**Abstract**

**Purpose:**

Typically, the current dose prediction models are limited to small amounts of data and require re-training for a specific site, often leading to suboptimal performance. We propose a site-agnostic, three dimensional dose distribution prediction model using deep learning that can leverage data from any treatment site, thus increasing the total data available to train the model. Applying our proposed model to a new target treatment site requires only a brief fine-tuning of the model to the new data and involves no modifications to the model input channels or its parameters. Thus, it can be efficiently adapted to a different treatment site, even with a small training dataset.

**Methods:**

This study uses two separate datasets/treatment sites: data from patients with prostate cancer treated with intensity-modulated radiation therapy (IMRT) (source data), and data from patients with head-and-neck cancer treated with volumetric modulated arc therapy (VMAT) (target data). We first developed a source model with 3D UNet architecture, trained from random initial weights on the source data. We evaluated the performance of this model on the source data. We then studied the generalizability of the model to the new target dataset via transfer learning. To do this, we built three more models, all with the same 3D UNet architecture: target model, adapted model, and combined model. The source and target models were trained on the source and target data from scratch/random initial weights, respectively. The adapted model fine-tuned the source model to the target domain by using the target data. Finally, the combined model was trained from random initial weights on a combined data pool consisting of both target and source datasets. We tested all four models on the target dataset and evaluated quantitative dose-volume-histogram (DVH) metrics for the planning target volume (PTV) and organs at risk (OARs).

**Results:**

When tested on the source treatment site, the source model accurately predicted the dose distributions with average (mean, max) absolute dose errors of (0.32%±0.14, 2.37%±0.93) (PTV) relative to the prescription dose, and highest mean dose error of 1.68%±0.76, and highest max dose error of 5.47%± 3.31 for femoral head right. The error in PTV dose coverage prediction is 3.21%±1.51 for $D_{98}$, 3.04%±1.69 for $D_{95}$ and 1.83%±1.01 for $D_{02}$. Averaging across all OARs, the source model predicted the OAR mean dose within 1.38% and the OAR max dose within 3.64%. For the target treatment site, the target model average (mean, max) absolute dose errors relative to the prescription dose for the PTV were (1.08%±0.95, 2.90%±1.35). Left cochlea had the highest mean and max dose errors of 5.37%±5.82 and 8.33%±8.88, respectively. The errors in PTV dose coverage prediction for $D_{98}$ and $D_{95}$ were 2.88%±1.59 and 2.55%±1.28, respectively. The target model can predict the OAR mean dose within 2.43% and the OAR max dose within 4.33% on average across all OARs.

**Conclusion:**

We developed a site-agnostic model for three dimensional dose prediction and tested its adaptability to a new target treatment site via transfer learning. Our proposed model can make accurate predictions with limited training data.

1.	Introduction

Radiation therapy (RT) is one of the main cancer treatment modalities. The advancement of novel RT modalities, such as intensity-modulated radiation therapy (IMRT) [1-7] and volume modulated arc therapy (VMAT) [8-13], has greatly improved treatment plan quality at the cost of increasing planning complexity and time. In radiation therapy treatment planning, the goal is to deliver the desired dose to the target while sparing sensitive nearby organs, known as organs at risk (OARs). The tradeoff between achieving these two goals makes treatment planning a complex process. Modern treatment planning techniques solve an inverse optimization problem [14]. The planner adjusts the hyperparameters of the fluence map optimization in a commercial treatment planning system (TPS) and tries to determine the optimal parameters to control the tradeoff between the plan objectives in a lengthy and tedious trial-and-error fashion. After this first phase, once the plan meets the planner's expectations, many rounds of plan review and consultation between the planner and the physician are required to ensure that the plan quality is clinically acceptable and meets the physician's desired tradeoffs/objectives. This is mainly because the physician's exact preferences for the tradeoff between target coverage and OAR sparing cannot easily and accurately be quantified and conveyed to the planner through the prescription. Moreover, some clinical objectives may be modified during planning, and these modifications may not have been addressed in the prescription.

Although the advancements in treatment planning systems have decreased human intervention and sped up the planning process, treatment planning is still not fully automated and is prone to variability in plan quality due to institutional guidelines, planner skills and planning preferences. Moreover, the process can still be tedious and lengthy (from hours to days), which can hinder the infusion of adaptive strategies and delay the delivery of the treatment, thus detracting from tumor local control and patient care. Another drawback of the current treatment planning systems is that they typically use dose-volume objectives. These objectives do not reflect the spatial variations of the dose within a structure, and they are blind to structures that aren't contoured. Developing a comprehensive automated planning tool requires eliminating the dependence on handcrafted features and providing spatial dose distributions. Recently, knowledge-based planning (KBP) methods have been developed to enhance the quality and efficiency of planning. KBP, of which Varian RapidPlan (Varian Medical Systems, Palo Alto, CA, USA) and Pinnacle Auto-Planning software (Philips Radiation Oncology, Fitchburg, WI, USA) are commercial examples, utilizes a large database of properly selected historical clinical plans to generate predictions for new patients [15-20]. The performance of KBP-based models depends on the size and diversity of the patient database and on the careful selection of the proper/most effective features extracted from the data. Before new advancements in deep learning (DL), traditional machine learning techniques were used in KBP. Thus, the application of KBP-based models is mostly limited to predicting small dimensional data, such as dose-volume histogram (DVH) or specific dosimetrist criteria, and it does not predict 3D dose distributions. The exception is the study by Shiraishi and Moore on KBP-based 3D dose distribution prediction in patients with prostate cancer [19]. In this study, instead of a simple functional parametrization, an artificial neural network (ANN) was incorporated into the KBP method to improve the KBP-based prediction techniques and predict achievable 3D dose distributions.

In recent years, artificial intelligence (AI) and deep learning have found wide applications in healthcare research. DL-based models are capable of learning their own features without human intervention and depend less on handcrafted features and domain knowledge. DL-based dose predictors can guide planners/dosimetrists to improve the plan quality and consistency during the plan optimization procedure. Several studies have addressed dose distribution prediction using deep learning on several sites such as

4prostate stereotactic body radiotherapy (SBRT) [21], prostate IMRT [22], prostate VMAT [23], head-and-neck IMRT [24-26], head-and-neck VMAT [27], and lung IMRT [28]. However, some challenges remain to be addressed for the successful deployment of AI in clinical practice [29].

In the radiotherapy treatment planning pipeline, the OARs that are considered and contoured vary from patient to patient and from one treatment site to another. However, the current dose prediction models require a fixed number of OARs as input and are typically site-specific. Thus, they need to be trained on the limited data available on a specific treatment site. Due to the difficulties in collecting medical data and the limited data available, the current site-specific dose predictor models may have overall limited performance. Also, applying a site-specific model to a new target treatment site with different OARs requires restructuring and retraining the model on the new target site with a different set of OARs.

In our proposed data preprocessing, each voxel on each OAR is ranked based on their distance to the PTV; on the basis of these distances, the voxels' corresponding dose values are then mapped onto them. Using a transfer learning technique, we demonstrate that our model performs accurately on a new target treatment site, even with limited training data. Applying our model to the new target treatment site does not require any modifications to the model parameters, or long re-training from random initial weights; rather, it requires only fine-tuning of the source model on the target domain.

Our dose predictor provides the planner with the 3D dose distribution as well as clinically applicable DVH metrics, so they can observe the predictions both visually and quantitatively prior to the planning process. This provides the means for physician/planners to include necessary tradeoffs in their decision making and to find a balance between desired and achievable objectives.

2.  Materials and Methods

2.A. Patient database

The dataset contains two subsets: 1) IMRT prostate and 2) VMAT head-and-neck cancer patient data. The prostate dataset includes 70 patients, which we randomly split into 54 training, 6 validation, and 10 test patients. We generated 1200 Pareto surface IMRT plans per patient by pseudo-randomizing the PTV and OAR tradeoff weights [30]. More details on the Pareto optimal plan are discussed in the next section. Since 1200 plans were generated for each patient, there were a total of 64,800 training, 7,200 validation, and 12,000 testing plans (Table 1). The head-and-neck dataset includes data from 58 patients, which we randomly split into 43 training, 5 validation, and 10 testing patients (Table 1). Due to inherent limitations with VMAT optimization, there is no fast optimizer to create VMAT plans in a similar way as we did for the prostate IMRT plans, so we directly exported head-and-neck VMAT plans from the Eclipse V13.7-V15.5 treatment planning system (Varian Medical Systems, Palo Alto, CA). The number of plans for different patients varies from 2-29. The average number of plans per patient is 11. These include approved plans, delivered plans, and intermediate plans that were not necessarily approved. The contoured OARs differ from one patient to another.

Table 1. Data split pattern for the two treatment sites.

|  | Training patients (plans) | Validation patients (plans) | Test patients (plans) | Total patients (plans) |
|---|---|---|---|---|
| Prostate dataset | 54 (64,800) | 6 (7,200) | 10 (12,000) | 70 (84,000) |
| Head-and-neck dataset | 43 (491) | 5 | 10 | 58 |

Data were acquired at 5 mm$^3$ voxel size for both prostate and head-and-neck cancer patients. For prostate cancer patients, contours for anatomical data consist of the PTV and the following organs-at-risk: body, bladder, rectum, left femur, and right femur. OARs for head-and-neck are body, left and right brachial plexus, brainstem, left and right cerebellum, left and right cochlea, constrictors, esophagus, larynx, mandible, left and right masseter, oral cavity, post arytenoid & cricoid space (PACS), left and right parotid, left and right submandibular gland (SMG), and spinal cord.

2.B. Pareto Optimal Plan Data for prostate cancer patients' data

Pareto optimal plans sample the Pareto surface. They are the solutions to a multi-criteria problem with various possible tradeoffs, particularly the tradeoffs between dose coverage of the target and dose sparing of the OARs. The goal is to deliver the prescribed dose to the PTV while minimizing the dose received by each OAR. The dose influence arrays are calculated and IMRT plans were generated that sampled the Pareto surface. Further details are described by Nguyen et al. [30]. In summary, the multi-criteria objective function includes the parameter $w_s$ $\forall s \in$ {PTV, OARs}. $w_s$ is the user-defined tradeoff structure weight for each objective function. Varying $w_s$ allows us to obtain different Pareto optimal solutions. The optimization problem was solved by using an in-house proximal-class first-order primal-dual algorithm, Chambolle-Pock [31], and pseudo-random plans were generated. A total of 1200 plans for each of the 70 patients were generated. The bounds for the controlled weights are such that the generated plans are highly likely to be within clinically acceptable bounds, though they may not necessarily be acceptable to a physician. Thus, the data include all kinds of tradeoffs, in terms of structure dose curves. We included all the plans during training and did not exclude lower quality plans, thus allowing the model to learn from a wide range of tradeoffs.

2.C. Data preprocessing and the deep learning neural network input

For IMRT prostate cancer patients, the PTV, OAR, and body masks are binary masks where each voxel value is equal to 1 if it is assigned to PTV/OAR/body, and 0 otherwise. For VMAT head-and-neck cancer patients, OAR and body masks are binary masks. However, since there is more than one prescription dose, the voxels in the PTV channel of the model input were defined to either contain the prescription dose value of their associated PTV, or 0 if the voxel did not belong to any of the PTVs. The data are normalized to a fixed value of 70 so that, similar to our prostate data, the maximum dose remains around 1. The voxel resolution of both contours and dose are 5 mm$^3$.

The input to the DL model is obtained from the structures' masks and the physician's desired DVH. In the data preprocessing step, first each voxel in each individual OAR is ranked based on its distance from the PTV. Then, based on these distance scores, dose values from the desired DVH are assigned to each voxel (Figure 1). This allows us to arrange all OARs in one channel. Thus, unlike the current deep learning dose predictors that allocate one input channel to each individual OAR, variations in the number of existing OARs will not affect the model input channels. So our proposed model can be applied to a new treatment site without needing to change the model parameters or the model structure, particularly the number of input channels.

The DL model takes three input channels: the mapped dose of the PTV, the mapped dose of the OARs, and the body binary mask.

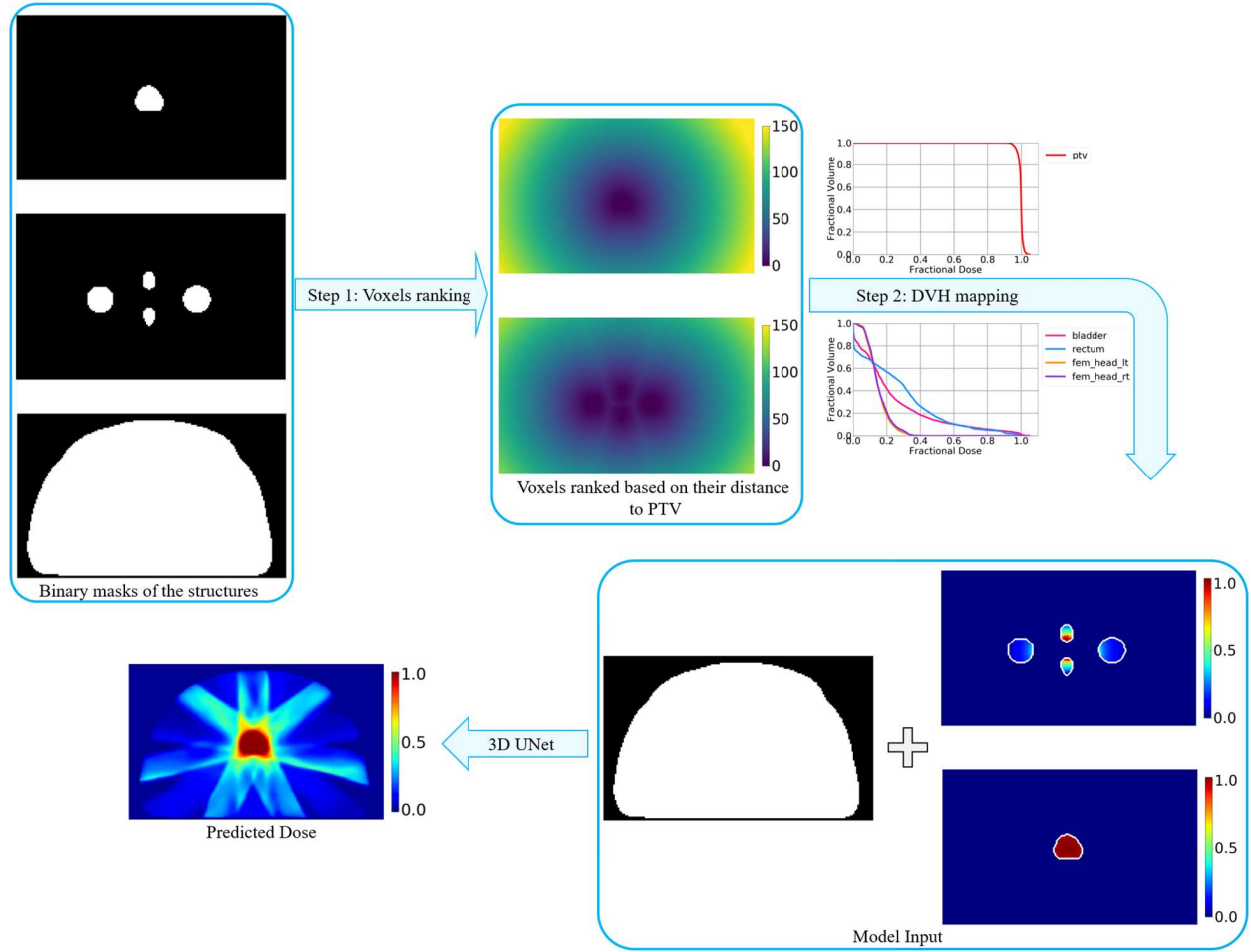

Figure 1. Data preprocessing. The distance-wise ranking of the structure voxels, followed by mapping the desired DVH onto the ranked structures. The mapped dose structures plus the body mask constitute the input to 3D UNet model.

The 3D arrays of mask and mapped doses are rearranged into 5D arrays to be fed into the model. Model channels take five dimensional tensors with the shape of ($n_{batch}$,x,y,z,$n_{channel}$). Data are cropped to a fixed size. This optimal kernel size is large enough to enclose all nonzero voxel values of the body, thus all the OARs. For prostate data, x=288, y=176, z=80; for head-and-neck data, x=160, y=160, z= 80. $n_{batch}$ =1 is the batch size, and $n_{channel}$ =3 is the number of channels. Thus, ($n_{batch}$,x,y,z,0) is the dose map array of the PTV, ($n_{batch}$,x,y,z,1) contains the dose map array of OARs, and ($n_{batch}$,x,y,z,2) is the body mask array.

2.D. Deep learning neural network architecture and parameters

We developed a 3D supervised convolutional neural network (CNN) based on the UNet [32] architecture. UNet is a CNN [33,34] that was initially designed for semantic segmentation of biomedical images and that can be used to predict a voxel-level spatial dose distribution. UNet incorporates both local and global features to make a pixel-wise prediction. The UNet architecture revolves around three main design ideas: 1) extracting the global features of the image via max pooling operators in the contraction half of the network, 2) returning the image to its original size via transposition operators or upsampling/deconvolution [35], and 3) propagating the local features from the contraction half of the UNet (downsampling arm) to the global information in the expansion half of the network (upsampling arm) via skip connections. Our 3D variant of the UNet model consists of the same three major parts: downsampling arm, bottleneck, and upsampling arm. It takes 3D volumetric inputs instead of the slice-by-slice input used in a 2D UNet. Thus, it predicts 3D volumetric dose distributions. The model consists of four max pooling operators [36] with a 2×2×2 pool between layers to reduce the feature size from the kernel size of 288×176×80 voxels down to 18×11×5 voxels for prostate data, and from 160×160×80 voxels down to 10×10×5 voxels for head-and-neck data, then back to their original size. A convolutional kernel of size 3×3×3 was implemented during the convolutions. Zero padding was used to maintain the feature size. The starting filter number was 16, with an expansion rate of 2. This increases the number of feature maps in the left half of the network by a factor of 2. The activation function in all the hidden layers is rectified linear units (ReLU) [37], which is defined as $f(x) = \max(0, x)$. The ReLU activation function improves the training process and the model performance. Compared to activation functions such as sigmoid or tanh, ReLU does not have problems with saturation or limited sensitivity. It improves sparsity, reduces the impact of vanishing gradients, and requires much fewer computational resources, thus speeding up the training process. The final activation is linear activation. A normalization operator updates the weights equally throughout the UNet and accelerates the convergence rate of the model. For more effective training, the convolution and ReLU operations are followed by Group Normalization [38]. Group Normalization is used as an alternative to Batch Normalization to avoid the Batch Normalization error due to inaccurate batch statistic estimations in small batch sizes. In our training, the batch size was equal to 1.

To prevent overfitting the model, we implemented dropout regularization [39]. The dropout rate for each layer of the model was calculated from [22]:

$$\text{layer dropout} = \text{dropout rate} \times \frac{\text{current number of filters}^{1/2}}{\text{max number of filters}}$$

We chose dropout rate = 0.1, starting filter number = 16, number of pool = 4, and expansion rate = 2. Thus, the max number of filters = $\text{round}[(\text{expansion rate}^{\text{number of pool}}) \times \text{starting filter number}] = 256$

For the first half of the UNet, in each layer:

$$\text{number of filters} = \text{round}[(\text{expansion rate}^{\text{layer number}-1}) \times \text{starting filter number}]$$

and for the second half of the UNet in each layer:

$$\text{number of filters} = \text{round}\left[\left(\text{expansion rate}^{2\times\frac{\text{number of layers}}{2}-\text{layer number}-1}\right) \times \text{starting filter number}\right]$$

The layer number takes values in the range of 1 to number of pool+1.

The Mean Squared Error (MSE) loss function is used to minimize the difference between the predicted dose and the ground truth dose. In prostate cancer patients, the ground truth dose is the optimized dose generated by the Chambolle-Pock algorithm; for head-and-neck cancer patients, the ground truth dose is the clinical plan exported from Eclipse TPS. From $MSE = \frac{1}{n}\sum_{i=1}^{n}(D_{predicted}^{i} - D_{ground\ truth}^{i})^2$, MSE loss is calculated in each i$^{th}$ voxel, summed over all n voxels. The Adam algorithm [40] was used as the optimizer to minimize the MSE loss function. The default Adam parameters are β$_1$=0.9 and β$_2$=0.999, decay=0, and the learning rate is 1e$^{-3}$. Learning rate is scheduled via ReduceLROnPlateau callback in Keras with a factor of 0.1 and minimum value of 1×10$^{-4}$, and patience=30. This means that if validation loss does not improve for 30 consecutive epochs, the learning rate is reduced by a factor of 10. We used early stopping to further prevent overfitting [41]. Early stopping is a regularization method that stops the network training when the validation loss does not improve for a set number of iterations/epochs. We terminated the training if the validation loss did not improve for 99 epochs. Based on training and validation loss values, we determined the optimal number of iterations at which the model converges enough and reaches a stable loss value while ensuring that overfitting doesn't happen. This value was 150k. The validation loss was checked every 10 iterations. To avoid recording a model that is overfitted to the data, the best model weights are saved at the end of the epoch at which the validation loss was the minimum. Finally, the instance of the model with the lowest validation loss was used as the best model to perform predictions and to evaluate the test data.

The model was built and implemented in Keras [42] with Tensorflow [43] as the back end. Training and testing were performed on an NVIDIA Tesla k80 and a V100 GPU with 12 and 32 GB RAM, respectively. The final network consists of 5 layers with 10,114,561 trainable parameters. Training for 150k iterations took 6 days on average, and prediction time was 4.12 seconds on a sample test patient on average.

2.E. Deep learning neural network performance evaluation

We refer to the prostate cancer patient dataset as the source data and the head-and-neck cancer patient dataset as the target data. We first developed the source model. This is a model trained from random initial weights on our source dataset: IMRT prostate cancer patient data. After ensuring the quality of the predictions made by the source model, we investigated the model's generalizability to a new treatment site via a transfer learning technique. To do this, we built three more models with the same 3D UNet architecture and hyperparameters: target model, adapted model, and combined model. The source and target models were trained from random initial weights on the source and target data, respectively. The adapted model fine-tuned the source model to a target domain by using the target data. Rather than random initial weights, the source mode weights serve as the initial weights for the adapted model. Thus, the adapted model requires much fewer training iterations before it converges (40k). Finally, the combined model was trained from random initial weights on a combined dataset consisting of both target and source data. We evaluated the performances of all four models on the target treatment site: VMAT head-and-neck cancer patient data. We evaluated the models' performances by comparing DVH metrics for both predicted dose distribution and ground truth dose distribution, and we calculated their corresponding errors as a percent of the prescription dose. DVH is commonly used as a plan evaluation tool to compare different treatment plans based on their corresponding radiation dose distributions within a volume of interest in a patient. D# is the dose received by at least #% of the volume of a specific structure. In other words, D# indicates that #% of the volume of that structure receives at least D# dose. So, for example, $D_{98}$ is the dose that 98% of the structure receives.

The calculated dose statistics include PTV mean and max dose (defined as $D_2$ by the ICRU-83 report [44]), PTV coverage ($D_{98}$, $D_{95}$, and $D_{02}$), and structure mean and max dose ($D_{mean}$ and $D_{max}$). Due to the variability in the number of PTV levels for head-and-neck cancer patients, we report the percentage errors as a percent of the highest prescription dose for the head-and-neck dataset.

3. Results

In the first step of our evaluation, we investigated the quality of the predictions made by the source model on the source site (prostate). Dose washes and the DVH plot are shown in Figure 2. OAR contours are drawn on the images. The absolute error in DVH metrics as the percentage of the prescription dose is shown for PTV and OARs.

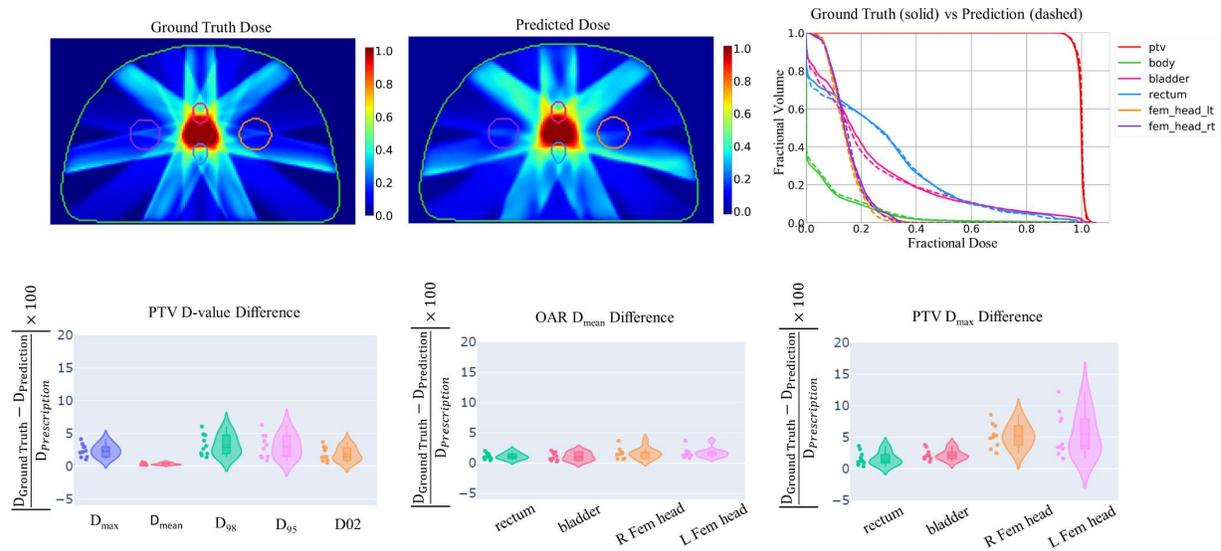

Figure 2. Performance evaluation of the source model on the source treatment site. (a) The ground truth and predicted dose distributions with the structures delineated, the DVH plot of the ground truth plan (solid lines) and the predicted dose (dotted lines), and (b) the violin plots of the error in DVH metrics in source model dose predictions.

Table 2**Error! Reference source not found.** shows the quantitative values of the errors in the source model's predictions on PTV coverage ($D_{95}$, $D_{98}$, and $D_{02}$) and PTV max and mean dose, as well as the percent error in OAR mean and max dose. The errors shown were found by $\frac{|D_{prediction} - D_{ground\ truth}|}{D_{Prescription}} \times 100$.

Table 2. Quantitative performance evaluation of the source model on the source treatment site. Percentage of absolute error in DVH metrics between the ground truth and predicted dose is shown for various regions of interest (ROI), consisting of the planning target volume (PTV), bladder, rectum, left and right femoral heads.

| ROI | DVH Metric | Base Model (%) |
|---|---|---|
| PTV | $D_{mean}$ | 0.32 ± 0.14 |
| | $D_{max}$ | 2.37 ± 0.93 |
| | $D_{98}$ | 3.21 ± 1.51 |
| | $D_{95}$ | 3.04 ± 1.69 |
| | $D_{02}$ | 1.83 ± 1.01 |
| Bladder | $D_{mean}$ | 1.13 ± 0.47 |
| | $D_{max}$ | 1.62 ± 1.01 |
| Rectum | $D_{mean}$ | 1.09 ± 0.63 |
|  | $D_{max}$ | 2.23 ± 0.80 |
| L Femoral Head | $D_{mean}$ | 1.61 ± 0.85 |
|  | $D_{max}$ | 5.24 ± 1.77 |
| R Femoral Head | $D_{mean}$ | 1.68 ± 0.76 |
|  | $D_{max}$ | 5.47 ± 3.31 |

The source model accurately predicted the dose distributions with average (mean, max) absolute dose errors of (0.32±0.14%, 2.37±0.93%) (PTV) relative to the prescription dose. The highest mean dose error was 1.68±0.76%, and the highest max dose error was 5.24±1.77% for femoral head left. The errors in PTV dose coverage prediction were 3.21±1.51% for $D_{98}$, 3.04±1.69% for $D_{95}$, and 1.83±1.01% for $D_{02}$. The source model predicted the OAR max dose within 1.38% and the OAR mean dose within 3.64%, when averaged across all OARs. Once we ensured the quality of predictions for the source model on prostate cancer patients' data, we developed three more models—target, adapted and combined models—and assessed their performance on the target treatment site: head-and-neck. The target and adapted models were trained on data from 43 head-and-neck cancer patients, for a total of 491 plans. The combined model was trained on the combination of head-and-neck and prostate cancer patient data. The comparison of the models' performance in terms of $D_{mean}$ for high impact OARs is shown in Figure 3. High impact OARs are the OARs that are most clinically relevant.

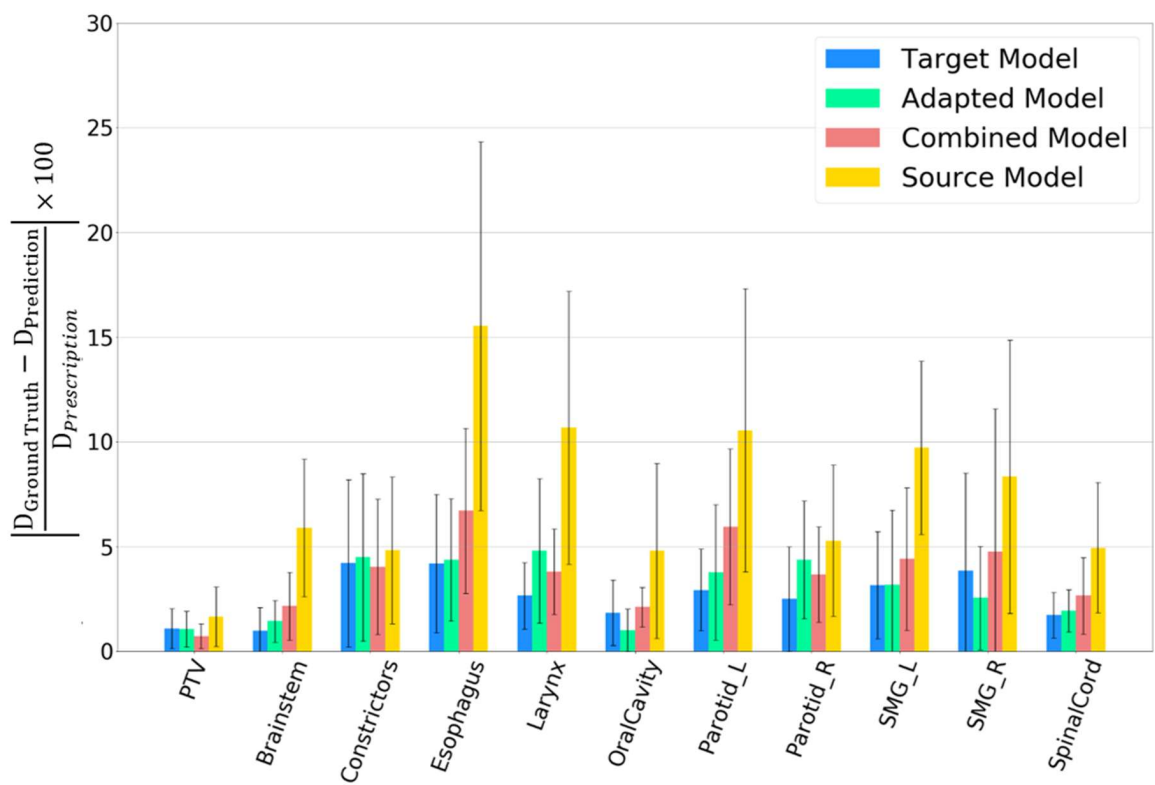

Figure 3. Performance comparison of adapted, target, combined, and source models on the target treatment site. The mean value and standard deviation of the $D_{mean}$ errors are shown for the planning target volume (PTV) and 10 high impact organs at risk consisting of brainstem, constrictors, esophagus, larynx, oral cavity, left and right parotids, left and right submodular glands, and spinal cord.

Table 3. Quantitative performance evaluation of the target, adapted, combined, and source models on the target treatment site. Percentage of absolute error in DVH metrics between the ground truth and predicted dose is shown for various regions of interest (ROI), consisting of the planning target volume (PTV) and organs at risk (OARs). OARs with high clinical impact consist of brainstem, constrictors, esophagus, larynx, oral cavity, left and right parotids, left and right submodular glands, and spinal cord. And OARs with low clinical impact consist of brachial plexus left and right, cerebellum left and right, cochlea left and right, mandible, masseter left and right, and post arytenoid & cricoid space (PACS). The structures with p-values > 0.05 are marked as well. Target-adapted (*), adapted-combined (●), and target-combined (♦).

|  | ROI | DVH Metric | Target model (%) | Adapted model (%) | Combined model (%) | Source model (%) |
|---|---|---|---|---|---|---|
|  | PTV | $D_{mean}$ | 1.08 ± 0.95 | 1.07 ± 0.85 | 0.73 ± 0.59 | 1.66 ± 1.42 |
|  |  | $D_{98}$ | 2.88 ± 1.59 | 2.24 ± 1.15 | 3.37 ± 1.99 | 24.34 ± 5.15 |
|  |  | $D_{95}$ | 2.55 ± 1.28 | 2.01 ± 0.92 | 2.69 ± 1.95 | 17.05 ± 4.12 |
|  |  | $D_{02}$ | 0.96 ± 0.60 | 0.66 ± 0.43 | 1.68 ± 1.95 | 1.51 ± 1.02 |
|  |  | $D_{max}$ | 2.90 ± 1.35 | 1.60 ± 1.09 | 4.34 ± 3.24 | 4.34 ± 3.24 |
| High Impact OARs | Brainstem | $D_{mean}$ | 0.98 ± 1.13♦ | 1.44 ± 0.99 | 2.16 ± 1.61♦ | 5.89 ± 3.28 |
|  |  | $D_{max}$ | 2.73 ± 1.89 | 2.79 ± 1.60 | 5.38 ± 4.04 | 9.66 ± 5.54 |
|  | Constrictors | $D_{mean}$ | 4.20 ± 3.99 | 4.50 ± 3.99 | 4.03 ± 3.24 | 4.82 ± 3.50 |
|  |  | $D_{max}$ | 4.47 ± 3.18 | 5.61 ± 4.44 | 3.86 ± 2.27 | 6.18 ± 4.73 |
|  | Esophagus | $D_{mean}$ | 4.18 ± 3.31♦ | 4.37 ± 2.92● | 6.71 ± 3.95♦,● | 15.53 ± 8.81 |
|  |  | $D_{max}$ | 5.04 ± 5.41 | 6.64 ± 4.75 | 8.68 ± 8.05 | 13.32 ± 7.37 |
|  | Larynx | $D_{mean}$ | 2.65 ± 1.58 | 4.80 ± 3.44 | 3.80 ± 2.04 | 10.68 ± 6.51 |
|  |  | $D_{max}$ | 4.58 ± 2.17 | 3.42 ± 1.85 | 4.80 ± 2.52 | 5.24 ± 3.98 |
|  | Oral cavity | $D_{mean}$ | 1.85 ± 1.57* | 1.0 ± 1.007* | 2.12 ± 0.94 | 4.80 ± 4.17 |
|  |  | $D_{max}$ | 5.95 ± 5.77 | 5.53 ± 4.88 | 4.24 ± 4.2 | 6.55 ± 6.60 |
|  | Parotid L | $D_{mean}$ | 2.93 ± 1.94♦ | 3.79 ± 3.23 | 5.94 ± 3.72♦ | 10.56 ± 6.75 |
|  |  | $D_{max}$ | 3.81 ± 5.10 | 3.47 ± 3.81 | 5.67 ± 5.40 | 8.49 ± 6.27 |
|  | Parotid R | $D_{mean}$ | 2.49 ± 2.50 | 4.37 ± 2.81 | 3.68 ± 2.26 | 5.28 ± 3.60 |
|  |  | $D_{max}$ | 4.09 ± 3.19 | 3.74 ± 3.04 | 3.56 ± 2.58 | 5.36 ± 3.90 |
|  | SMG L | $D_{mean}$ | 3.16 ± 2.56 | 3.19 ± 3.55 | 4.41 ± 3.4 | 9.72 ± 4.13 |
|  |  | $D_{max}$ | 3.43 ± 1.83 | 3.42 ± 2.05 | 4.86 ± 5.74 | 6.44 ± 4.89 |
|  | SMG R | $D_{mean}$ | 3.86 ± 4.64 | 2.55 ± 2.48 | 8.14 ± 10.44 | 8.34 ± 6.52 |
|  |  | $D_{max}$ | 3.80 ± 2.24 | 3.22 ± 1.92 | 3.13 ± 2.83 | 3.63 ± 2.65 |
|  | Spinal cord | $D_{mean}$ | 1.74 ± 1.09 | 1.93 ± 1.00 | 2.65 ± 1.81 | 4.9 ± 3.11 |
|  |  | $D_{max}$ | 2.55 ± 1.25 | 3.72 ± 2.60 | 6.72 ± 2.60 | 8.69 ± 4.33 |
| Low Impact OARs | Body | $D_{mean}$ | 0.82 ± 0.93 | 0.76 ± 0.99 | 0.97 ± 0.94 | 2.56 ± 1.85 |
|  |  | $D_{max}$ | 2.87 ± 1.38 | 1.62 ± 1.08 | 2.49 ± 1.96 | 4.29 ± 3.22 |
|  | BrachialPlexus L | $D_{mean}$ | 1.47 ± 0.89 | 2.71 ± 1.25 | 2.61 ± 2.34 | 6.58 ± 3.88 |
|  |  | $D_{max}$ | 1.01 ± 0.50 | 1.06 ± 0.72 | 1.19 ± 0.63 | 6.83 ± 3.20 |
|  | BrachialPlexus R | $D_{mean}$ | 2.54 ± 2.17 | 2.30 ± 2.09 | 2.16 ± 1.23 | 10.09 ± 3.12 |
|  |  | $D_{max}$ | 2.50 ± 1.94 | 2.77 ± 2.06 | 2.12 ± 1.62 | 6.52 ± 4.01 |
|  | Cerebellum L | $D_{mean}$ | 1.97 ± 2.10 | 2.50 ± 1.70 | 3.26 ± 2.00 | 6.56 ± 5.39 |
|  |  | $D_{max}$ | 6.74 ± 4.90 | 7.58 ± 5.06 | 6.44 ± 3.39 | 14.71 ± 7.36 |
|  | Cerebellum R | $D_{mean}$ | 1.66 ± 1.17 | 1.34 ± 1.17 | 2.23 ± 2.42 | 6.53 ± 4.51 |
|  |  | $D_{max}$ | 5.39 ± 3.81 | 3.37 ± 2.28 | 5.66 ± 4.49 | 10.86 ± 5.96 |
|  | Cochlea L | $D_{mean}$ | 5.37 ± 5.82 | 5.10 ± 6.63 | 6.15 ± 5.97 | 8.95 ± 7.06 |
| | | $D_{max}$ | 8.33 ± 8.88 | 7.84 ± 9.11 | 8.39 ± 8.50 | 12.89 ± 12.22 |
| | Cochlea R | $D_{mean}$ | 3.02 ± 3.73 | 3.29 ± 3.35 | 3.73 ± 4.64 | 7.79 ± 5.73 |
| | | $D_{max}$ | 3.65 ± 4.67 | 4.02 ± 4.09 | 4.63 ± 5.54 | 10.47 ± 9.36 |
| | Mandible | $D_{mean}$ | 1.49 ± 0.78 | 1.37 ± 0.69 | 2.02 ± 1.23 | 2.32 ± 1.65 |
| | | $D_{max}$ | 3.18 ± 2.14 | 3.29 ± 2.63 | 3.26 ± 3.78 | 5.33 ± 4.11 |
| | Masseter L | $D_{mean}$ | 2.76 ± 2.03 | 2.35 ± 2.01 | 2.92 ± 1.76 | 9.18 ± 6.32 |
| | | $D_{max}$ | 3.80 ± 4.23 | 3.57 ± 3.46 | 5.67 ± 4.41 | 7.91 ± 3.95 |
| | Masseter R | $D_{mean}$ | 4.10 ± 1.42 | 3.49 ± 1.49● | 6.12 ± 2.64● | 11.19 ± 5.31 |
| | | $D_{max}$ | 4.73 ± 2.60 | 5.10 ± 2.69 | 7.21 ± 6.24 | 13.93 ± 8.76 |
| | PACS | $D_{mean}$ | 3.58 ± 2.92 | 4.94 ± 3.48 | 5.57 ± 4.49 | 16.05 ± 9.12 |
| | | $D_{max}$ | 5.44 ± 3.56 | 7.37 ± 3.48 | 6.35 ± 3.47 | 8.02 ± 4.22 |

Table 3 Shows the errors in DVH metrics for the PTV and 21 OARs, split into high and low impact OARs. $D_{98}$, $D_{95}$, $D_{02}$, $D_{mean}$ and $D_{max}$ values are reported for PTV. And $D_{mean}$, $D_{max}$ are reported for OARs. P-values were calculated from a paired two-sided t-test for each pair of models. We chose a statistical significance threshold (alpha) of 0.05. Comparisons that were not statistically significant (p-values >0.05) are marked**Error! Reference source not found.**.

From Table 3 and Figure 3, the performance of the source model on the new dataset was suboptimal. To keep the table less crowded, the p-values for the comparison between the source model and the other three models are not shown. For a large number of OARs (>40%), the predictions from the source model were significantly different from the predictions from the other three models. This was expected, since the model was trained on one dataset and tested on an unseen dataset. The adapted, target and combined models had relatively comparable performance.

We further studied the effect of the training size on the models' performance by evaluating the MSE loss between the predicted and the ground truth dose in 10% isodose volume for the three top models (target, adapted and combined models) as a function of training size. v% isodose volume is defined as the region/volume that receives dose values greater than or equal to v% of the prescription dose value. We took 10% isodose volume for the ground truth dose and then calculated the MSE in that volume between the ground truth and the predicted volumes. MSE is calculated from $\text{MSE} = \frac{1}{n}\sum_{i=1}^{n}(\text{Dose}^i_{predicted} - \text{Dose}^i_{ground\ truth})^2$ in each body voxel, i, summed over all n voxels. 1, 4, 10, 14, 23, 32 patients were chosen randomly from the pool of training data to construct the training datasets. We chose these patients from those with eight plans or more. For each patient, we randomly selected only eight plans per patient to allow a fair comparison of the performance. The validation dataset included five patients and was kept the same among all different training sizes. At each training size, we repeated the training 10 times to find the mean and the standard deviation. Models were tested on a test set of size 10 (Figure 4).

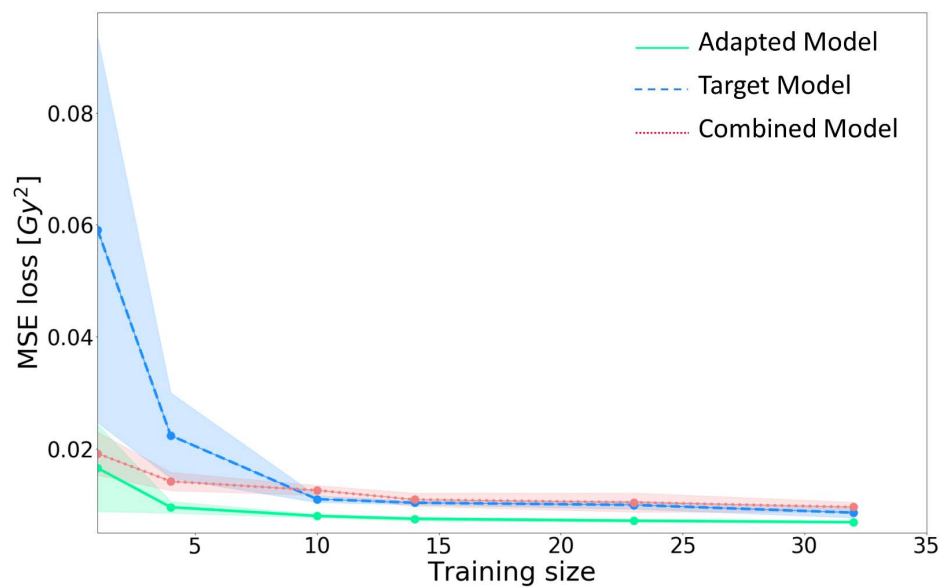

Figure 4. Mean squared error (MSE) loss comparison between the predicted and ground truth doses in 10% isodose volume for the adapted, target and combined models as a function of training size. Eight plans per patient were randomly chosen.

The results (Figure 4) show the optimal performance of the adapted model (green solid line) on small training data sizes. Combined model (red dotted line) had close performance to the adapted model. Target model (blue dashed line) does not perform optimal on small sizes. For larger training sizes all models perform comparably.

To check the possible effect of the data imbalance and the model's performance gain upon adding data from the target dataset, we further looked into the performance of the combined model when tested on the source dataset. The results are shown in Table 4. The performance of the combined model on the source dataset is comparable to the performance of the source model on the source dataset and for some OARs (shown in bold font) performance has improved.

Table 4. The performance of the combined model on the source dataset. The errors in DVH metrics for the regions of interest (ROI) consisting of the planning target volume (PTV) and organs at risk (OARs) are shown. OAR list consist of the planning target volume (PTV), bladder, rectum, left and right femoral heads. Bold font OARs indicate improvement in the model performance over the source model predictions on the source data.

| ROI | DVH Metric | Base Model (%) |
|---|---|---|
| PTV | $D_{max}$ | 3.02 ± 0.77 |
|  | $D_{mean}$ | **0.27 ± 0.18** |
|  | $D_{98}$ | 3.44 ± 1.34 |
|  | $D_{95}$ | 3.7 ± 1.36 |
|  | $D_{02}$ | 2.58 ± 0.98 |
| Bladder | $D_{mean}$ | **0.91 ± 0.37** |
|  | $D_{max}$ | **1.39 ± 0.78** |
| Rectum | $D_{mean}$ | 0.75 ± 0.57 |
|  | $D_{max}$ | 2.77 ± 0.85 |
| L Femoral Head | $D_{mean}$ | **1.29 ± 1.10** |
| | $D_{max}$ | 5.29 ± 1.90 |
| R Femoral Head | $D_{mean}$ | **1.33 ± 0.75** |
| | $D_{max}$ | 5.55 ± 4.29 |

4. Discussion

Using the supervised learning technique, we developed a site-agnostic deep neural network for 3D radiotherapy dose distribution prediction that addresses some existing limitations of the current DL-based dose distribution prediction models. Current deep learning–based dose distribution prediction models are inflexible with respect to adding more OARs and are typically site-dependent. A model that is trained to have a number of specific structures as inputs, can handle dose prediction on patients lacking some of the structures if the data augmentation and training was handled properly to account for missing structures. However, model cannot account for new structures without entirely retraining the model. The study done by Nguyen et al. shows an example of this limitation [27]. Similarly, a site-specific model cannot be applied to a different treatment site and the model needs to be restructured accordingly and retrained on the new treatment site with a different set of OARs. Our proposed preprocessing includes ranking each voxel in each structure based on its distance to the PTV. Then, the dose values from desired DVHs are mapped onto each voxel based on these distance ranks (Figure 1). With this method, we arrange all OAR information in one input channel. Thus, the same model can be applied to any treatment site without needing to alter the model's input channels.

We developed four models in this study and to reduce the variability in the study, we kept the hyperparameters the same across all four models. Models are trained to learn the mapping of the ground truth radiation dose distribution from the body binary masks and the PTV and OAR mapped dose masks. Dose predictions are then made for test patients based solely on their PTV and OAR masks. In the first phase, we developed a source model, i.e. the model trained on source treatment site data from random initial weights. The dose distributions and DVH comparisons show the source model's accurate performance on the source treatment site data (Figure 2, Table 2). The average absolute dose errors for $D_{mean}$ and $D_{max}$ PTV over all test patients were 0.32% and 2.37%, respectively. Averaging across all OARs, the model predicted the OAR mean dose within 1.38% and the OAR max dose within 3.64%.

Once we were sure of the quality of the predictions, we then used the transfer learning method to evaluate the generalizability and performance of the model on the target treatment site. Generalizability is an essential consideration for developing robust deep learning models that can perform optimally on new unseen target datasets. We developed the target, adapted and combined models and tested the performance of the models on the target dataset. We found that the target and adapted models' performances were similar and only showed statistically significant performance differences on predicted dose values for the oral cavity (Figure 3, Table 3). As expected, the source model trained on the source dataset performed suboptimally on the new unseen test data (Figure 3, Table 3). Although the source model had the poorest performance of the four models, it produced reasonable dose estimations on a previously unseen treatment site. The average of the $D_{mean}$ and $D_{max}$ values for OARs lies within 2.32% and 16.05%, and 3.63% and 14.71%, respectively.

One limitation of the current model pertains to the availability of training data. Collecting clinical data is often a difficult task, and there may be insufficient data available to train the model. This can limit the

15model's performance. Our proposed model can leverage data from any treatment site, thus increasing the total data available to train the model. As seen from the results (Figure 3, Table 3), compared to the source model (trained solely on the source dataset), the combined model made accurate predictions on the source dataset (Table 4) and predicted lower error values on $D_{mean}$ for PTV, bladder, left and right femoral heads. This can reflect the performance enhancement that resulted from increasing the total data available to train the model. However, the combined model did not perform as well as the target and adapted models on the target site. The combined model was trained on the combined datasets of two different modalities and two different treatment sites: prostate IMRT and head-and-neck VMAT. Furthermore, the data pool is largely imbalanced since the total number of plans for the target site was much smaller than for the source dataset (491 vs. 64800). This could explain the suboptimal performance of the combined model on the target dataset.

We also tested the dependence of the three top models' performance on the training size. We calculated the MSE loss between the ground truth and the predicted dose on a 10% isodose volume of the ground truth at each training size, and we showed that the model's performance depends on the size of the data pool it has been trained on. The adapted model, shown in green solid line in Figure 4, had lower loss values than the other two models, even for limited target data sizes. The adapted model does not need much data to learn the mapping on the new data set. The combined model performed second best ad performs better than the target model. This could be because the combined model has more data available for training than the target model. The good performance of the adapted and combined models is especially beneficial in cases where limited data are available for training a model on a new dataset. The model trained on a different treatment site with possibly enough training data available can be adapted to the new target domain data with limited data available (adapted model). Or the model can be trained on a pool of data for several treatment sites (combined model). The target model had the poorest performance. The target model did not seem to have seen enough data during the training to effectively learn the mapping between the ground truth and the predictions. Being trained from random initial weights without enough training data available may have contributed to the poor performance of the target model with small training data size. With larger training sizes, the three models converged and had similar MSE loss values. The maximum number of patients in the Figure 4 plot is 32 (256 plans).

Another important feature of our model is that it incorporates the desired DVH information into the input. Recent research has aimed to enhance planning quality and consistency and to eliminate the iterative process by developing easy to use tools for physicians and planners. Multi-criteria optimization (MCO), of which the RayStation multi-criteria optimization algorithm (RaySearch Laboratories AB, Stockholm, Sweden) is an example, facilitates exploring the tradeoffs and optimization of the tradeoffs for the user (physician/planner) in IMRT and VMAT [45-51]. There are some limitations with MCO. For example, it still requires some manual input, such as structure weight tuning, beam geometry tuning, selecting proper dose volume constraints, and tradeoffs between PTV coverage and OAR sparing. Moreover, some dose deviations in the target and critical structures can occur due to the conversion of navigated treatment plans to a deliverable machine setting [49]. Thus, the initially acceptable plan may not be acceptable for clinical delivery. Thus, to further automate the dose prediction pipeline, it would be valuable to have a model that takes dose volume constraints as its input. To our knowledge, our model is the first site-agnostic DL-based radiotherapy dose predictor that incorporates DVH information and is generalizable to different treatment sites. Since the DVH information is incorporated in the input data, our proposed model has the potential to make predictions based on clinical dose constraints and tradeoffs, meaning it can be tunable based on different prescriptions. Incorporating tunable DVHs and being able to control the tradeoffs would allow for an interactive, real-time tradeoff navigation and a fast dose prediction procedure. The desired DVH that our model takes as input can come from a dose predictor model, which doesn't necessarily require a DVH as input. Or it can come from a treatment planning system, possibly after including the planner's or physician's dose preferences and tweaks, i.e., the fine-tuned DVH. The next step in expanding the application of our model would be to feed this desired DVH to our model to obtain a patient-specific dose prediction. Moreover, the current version of the model takes feasible plans (DVHs). However, the desired DVH can be infeasible. Another improvement to our current model would be to incorporate infeasible DVHs and have the network project these back to the feasible space.

5. Conclusion

In this study, we developed multiple models to assess the performance of and generalizability of our proposed framework. We included two separate datasets/treatment sites: data from patients with prostate cancer treated with intensity-modulated radiation therapy (IMRT) (source data), and data from patients with head-and-neck cancer treated with volumetric modulated arc therapy (VMAT) (target data). We then developed 4 different models based on various training schemes and dataset combinations—source model, target model, adapted model, and combined model—and evaluated their performance and generalizability. Our proposed framework can incorporate physician preferences into the dose prediction process, it has good performance in small training data sizes, and it is independent of the treatment site. Being applicable to different sites makes this model a universal tool for faster dose prediction.

6. Acknowledgements:

This study was supported by the NIH Grant No. R01CA237269. The authors thank Dr. Jonathan Feinberg for editing the manuscript.

17REFERENCES

1. Brahme A. Optimization of stationary and moving beam radiation therapy techniques. *Radiotherapy and Oncology.* 1988;12(2):129-140.
2. Bortfeld T, Bürkelbach J, Boesecke R, Schlegel W. Methods of image reconstruction from projections applied to conformation radiotherapy. *Physics in Medicine & Biology.* 1990;35(10):1423.
3. Bortfeld TR, Kahler DL, Waldron TJ, Boyer AL. X-ray field compensation with multileaf collimators. *International Journal of Radiation Oncology• Biology• Physics.* 1994;28(3):723-730.
4. Convery D, Rosenbloom M. The generation of intensity-modulated fields for conformal radiotherapy by dynamic collimation. *Physics in medicine & biology.* 1992;37(6):1359.
5. Xia P, Verhey LJ. Multileaf collimator leaf sequencing algorithm for intensity modulated beams with multiple static segments. *Medical Physics.* 1998;25(8):1424-1434.
6. Keller-Reichenbecher M-A, Bortfeld T, Levegrün S, Stein J, Preiser K, Schlegel W. Intensity modulation with the "step and shoot" technique using a commercial MLC: A planning study. *International Journal of Radiation Oncology\* Biology\* Physics.* 1999;45(5):1315-1324.
7. Webb S. Optimisation of conformal radiotherapy dose distribution by simulated annealing. *Physics in Medicine & Biology.* 1989;34(10):1349.
8. Yu CX. Intensity-modulated arc therapy with dynamic multileaf collimation: an alternative to tomotherapy. *Physics in Medicine & Biology.* 1995;40(9):1435.
9. Otto K. Volumetric modulated arc therapy: IMRT in a single gantry arc. *Medical physics.* 2008;35(1):310-317.
10. Earl M, Shepard D, Naqvi S, Li X, Yu C. Inverse planning for intensity-modulated arc therapy using direct aperture optimization. *Physics in Medicine & Biology.* 2003;48(8):1075.
11. Palma D, Vollans E, James K, et al. Volumetric modulated arc therapy for delivery of prostate radiotherapy: comparison with intensity-modulated radiotherapy and three-dimensional conformal radiotherapy. *International Journal of Radiation Oncology\* Biology\* Physics.* 2008;72(4):996-1001.
12. Shaffer R, Nichol AM, Vollans E, et al. A comparison of volumetric modulated arc therapy and conventional intensity-modulated radiotherapy for frontal and temporal high-grade gliomas. *International Journal of Radiation Oncology\* Biology\* Physics.* 2010;76(4):1177-1184.
13. Shaffer R, Morris W, Moiseenko V, et al. Volumetric modulated Arc therapy and conventional intensity-modulated radiotherapy for simultaneous maximal intraprostatic boost: a planning comparison study. *Clinical oncology.* 2009;21(5):401-407.
14. Oelfke U, Bortfeld T. Inverse planning for photon and proton beams. *Medical dosimetry.* 2001;26(2):113-124.
15. Ge Y, Wu QJ. Knowledge-based planning for intensity-modulated radiation therapy: a review of data-driven approaches. *Medical physics.* 2019;46(6):2760-2775.
16. Kamima T, Ueda Y, Fukunaga J-i, et al. Multi-institutional evaluation of knowledge-based planning performance of volumetric modulated arc therapy (VMAT) for head and neck cancer. *Physica Medica.* 2019;64:174-181.
17. Hussein M, South CP, Barry MA, et al. Clinical validation and benchmarking of knowledge-based IMRT and VMAT treatment planning in pelvic anatomy. *Radiotherapy and Oncology.* 2016;120(3):473-479.